\documentclass[final,authoryear,5p,times,twocolumn]{elsarticle}

\usepackage{graphicx}
\usepackage[tight,footnotesize]{subfigure}

\usepackage{amssymb}
\usepackage{amsmath}

\usepackage[nodots]{numcompress}

\journal{Journal of Neuroscience Methods}

\begin{document}

\begin{frontmatter}



\title{Multivariate Temporal Dictionary Learning for EEG}

\author[label1,label2]{Q. Barth\'elemy}
\author[label1]{C. Gouy-Pailler}
\author[label1]{Y. Isaac}
\author[label1]{A. Souloumiac}
\author[label1]{A. Larue \corref{cor1}}
\author[label1,label2]{J.I. Mars}
\address[label1]{CEA, LIST, Data Analysis Tools Laboratory, Gif-sur-Yvette Cedex, 91191, France}
\address[label2]{GIPSA-lab, DIS, UMR 5216 CNRS, Grenoble INP, Grenoble, 38402, France}

\cortext[cor1]{Corresponding author. CEA, LIST, Data Analysis Tools Laboratory, Bat DIGITEO 565 PC 192, Gif-sur-Yvette Cedex, 91191, France. Tel: + 33 1 69 08 84 39. }
\ead{ quentin.barthelemy@cea.fr, cedric.gouy-pailler@cea.fr, yoann.isaac@cea.fr, antoine.souloumiac@cea.fr, anthony.larue@cea.fr, jerome.mars@gipsa-lab.grenoble-inp.fr }


\begin{abstract}
This article addresses the issue of representing electroencephalographic (EEG) signals in an efficient way. While classical approaches use a fixed Gabor dictionary to analyze EEG signals, this article proposes a data-driven method to obtain an adapted dictionary. To reach an efficient dictionary learning, appropriate spatial and temporal modeling is required. Inter-channels links are taken into account in the spatial multivariate model, and shift-invariance is used for the temporal model. Multivariate learned kernels are informative (a few atoms code plentiful energy) and interpretable (the atoms can have a physiological meaning). Using real EEG data, the proposed method is shown to outperform the classical multichannel matching pursuit used with a Gabor dictionary, as measured by the representative power of the learned dictionary and its spatial flexibility. Moreover, dictionary learning can capture interpretable patterns: this ability is illustrated on real data, learning a P300 evoked potential.
\end{abstract}


\end{frontmatter}



\section{Introduction}
\label{sec:intro}

Scalp electroencephalography (EEG) measures electrical activity produced by post-synaptic potentials of large neuronal assemblies. Although this old medical imaging technique suffers from poor spatial resolution, EEG is still widely used in medical contexts (\textit{e.g.} sleep analysis, anesthesia and coma monitoring, encephalopathies) as well as entertainment and rehabilitation contexts (Brain-Computer Interfaces -- BCI). EEG devices are relatively cheap compared to other imaging techniques (\textit{e.g.} MEG, fMRI, PET), and they offer both high temporal resolution (a short period of time between two acquisitions) and very low latency (a delay between the mental task and the recording on the electrodes). 

These features are of particular concern for the practitioner interested in \citep{Sanei2007}:
\begin{itemize} \itemsep-3pt
	\item Event-related potentials (ERPs) or evoked potentials: transient electrical activity that results from external sensory stimulation (\textit{e.g.} P300); 
	\item Steady-state evoked potentials: oscillatory brain activity that results from repetitive external sensory stimulation; 
	\item Event-related synchronizations/desynchronizations (ERS/ERD): oscillatory activity that results from involvement of a specialized part of the brain; for example, activation of the primary motor area, known as $\mu$ ($8$--$13$~Hz) or $\beta$ ($13$--$30$~Hz) bandpower synchronization or desynchronization, have been widely studied; 
	\item Epileptic activity: transient electrical bursts of parts of the brain.
\end{itemize}
While EEG devices are known to be able to record such aforementioned activities through the wide areas of the sensors, methodologists are usually necessary in EEG experiments. Indeed, they have to provide the practitioners with tools that can capture the temporal, frequential and spatial content of an EEG. Consequently, signal interpretation usually yields a \textit{representation} problem: which dictionary is able to best represent the information recorded in the EEG?

Fourier and wavelets dictionaries allow spectral analysis of the signals through well-defined mathematical bases \citep{Durka2007,Mallat2009}, although they show a lack of flexibility to represent the shape diversity of EEG patterns. The Gabor dictionary has also attracted high interest due to its temporal shift-invariance property. Nevertheless, it also suffers from a lack of flexibility to represent evoked potentials and EEG bursts \citep{Niedermeyer2004}. For example widely studied sleep activities such as spindles (centroparietal or frontal areas) consist of a complex EEG shape; as same epileptic activities such as inter-epileptic peaks are another examples of repeatable and complexely shaped cerebral activities \citep{Niedermeyer2004}. In these two cases practitioners should probably benefit from a custom-based dictionary approach over Fourier or wavelets dictionaries. While these approaches are based on \textit{a priori} models of the data, recent methodological developments focus on data-driven representations: \textit{dictionary learning} algorithms \citep{Tosic2011}. 

In EEG analysis, the spatial modeling consists of taking into account inter-channels links, and this has been done in several studies searching for more spatial flexibility \citep{Durka2007}. The EEG temporal modeling is more difficult. Some approaches use a hypothesis of temporal stationarity and treat only the spatial aspect, but this brings about loss of information. Other approaches use the generic Gabor dictionary which is shift-invariant. But it remains difficult to learn an EEG dictionary that integrates these two aspects \citep{Jost2005,Hamner2011}.

In this article, time-frequency analysis tools that are used to deal with EEG are first reviewed in Section \ref{sec:state_of_the_art}.
Then, temporal modeling is proposed based on the shift-invariance, and a spatial model called multivariate to provide an efficient dictionary learning in Section \ref{sec:multivariate_approach}.
As validation, the multivariate methods are applied to real EEG data, and then compared to other methods in Section \ref{sec:exp1}. Finally, to show the interpretability of the learned kernels, methods are applied for learning the P300 evoked potential.


\section{EEG analysis}
\label{sec:state_of_the_art}

In this section, some of the classical signal processing tools that are applied to EEG data for time-frequency analysis are reviewed: monochannnel and multichannel sparse approximations, shift-invariant dictionaries, and dictionary learning algorithms.

\subsection{Monochannel sparse approximation}
\label{ssec:mono_sparse_app}

In this paragraph, the EEG analysis is provided independently for the different electrodes / channels, so it is called monochannel.
A single channel signal $y \in \mathbb{R}^N$ of $N$ temporal samples and a normalized dictionary $\Phi \in \mathbb{R}^{N \times M}$ composed of $M$ time-frequency atoms $\left\{ \phi_m \right\}^{M}_{m=1}$ are considered.
The monochannel decomposition of the signal $y$ is carried out on the dictionary $\Phi$ such that: 
\begin{equation} 
	\label{eq:model}
	y = \Phi x + \epsilon \: ,
\end{equation}
assuming $x \in \mathbb{R}^M$ for the coding coefficients, and $\epsilon \in \mathbb{R}^N$ for the residual error. The approximation $\hat{y}$ is $\Phi x$. 
The dictionary is redundant since $M > N$, and thus the linear system of Eq.~(\ref{eq:model}) is under-determined. Consequently, sparsity, smoothness or another constraint is needed to regularize the solution $x$.
Considering the constant $K \ll M$, the sparse approximation is written as:
\begin{equation} 
	\label{eq:approx_parcimonieuse}
	\text{min}_x \left\| \: y - \Phi x \: \right\|^2 \ \text{s.t.} \ \left\| x \right\|_0 \! \leq \! K \: ,
\end{equation}
with $\left\| . \right\| $ for the Frobenius norm, and $\left\| x \right\|_0$ for the $\ell_0$ pseudo-norm, defined as the number of nonzero elements of vector $x$.
The well-known Matching Pursuit (MP) \citep{Mallat1993a} tackles this difficult problem iteratively, but in a suboptimal way. 
At iteration $k$, it iteratively selects the atom that is the most correlated to the residue $\epsilon^{k-1}$ as\footnote{$\left\langle A,B \right\rangle = \text{Tr} (B^H A)$ is the matrix scalar product.}
:
\begin{equation}
	\label{eq:selection0}
	m^k = \text{arg} \; \text{max}_m \left| \;\left\langle \epsilon^{k-1}, \phi_m \right\rangle \; \right| \: .
\end{equation} 
To carry out time-frequency analysis for EEG, MP is applied \citep{Durka2001} with the Gabor parametric dictionary, which is a generic dictionary that is widely used to study EEG signals.

\subsection{Multichannel sparse approximations}
\label{ssec:mult_sparse_app}

Hereafter, the EEG signal $y \in \mathbb{R}^{N \times C}$ composed of several channels $c=1..C$ are considered. The $c^{th}$ channel of the signal $y$ is denoted by $y[c]$.
The following reviewed methods link these channels spatially with the multichannel model illustrated in Fig.~\ref{fig:models}(left).

A multichannel MP \citep{Gribonval2003} was set up using a spatial (or topographic) prior based on structured sparsity. This model is an extension of Eq.~(\ref{eq:model}), with $y \in \mathbb{R}^{N \times C}$, $\Phi \in \mathbb{R}^{N \times M}$ and $x \in \mathbb{R}^{M \times C}$.
The multichannel model linearly mixes an atom in the channels, each channel being characterized by a coefficient. The underlying assumption is that few EEG events are spatially spread over in all of the different channels.

In \citep{Durka2007}, different multichannel selections are enumerated:
\begin{itemize} \itemsep-3pt
  \item the original multichannel MP \citep{Gribonval2003}, called MMP\_1 by Durka, selects the maximal energy such that:
  		\begin{equation}
  			\label{eq:selection1}
				m^k = \text{arg} \; \text{max}_m \sum_{c=1}^C \left| \; \left\langle \epsilon^{k-1}[c], \phi_m \right\rangle \; \right|^2 \: ;
			\end{equation} 
  \item the multichannel MP \citep{Durka2005} called MMP\_2 selects the maximal multichannel scalar product (also called average correlation):
    	\begin{equation}
    		\label{eq:selection2}
				m^k = \text{arg} \; \text{max}_m \left| \; \sum_{c=1}^C  \left\langle \epsilon^{k-1}[c], \phi_m \right\rangle \; \right| \: .
			\end{equation} 
			Due to absolute values, selection (\ref{eq:selection1}) allows the atom $\phi_m$ to be in-phase or in opposite phase with the component $\epsilon^{k-1}[c]$,
			contrary to the more constraining selection (\ref{eq:selection2}) which prefers $\phi_m$ to be in-phase with $\epsilon^{k-1}[c]$ (thus giving the same polarities across channels).
  \item the multichannel MP \citep{Matysiak2005} called MMP\_3 selects the maximal energy as Eq.~(\ref{eq:selection1}), but with complex coefficients that allows it to have varying phases: each channel $c$ has its own phase $\varphi_c$, as also studied in \citep{Gratkowski2007};
	\item the multichannel MP \citep{Sieluzycki2009b}, which will call MMP\_4 selects the maximal multichannel scalar product as Eq.~(\ref{eq:selection2}), with complex coefficients that allows it to have a phase $\varphi_c$ for each channel.
\end{itemize}

Note that other algorithms deal with EEG decomposition. For example, a multichannel decomposition was proposed in \citep{Koenig2001}, but it was based on the method of frames \citep{Daubechies1988}.

\subsection{Shift-invariant dictionaries}

The dictionary $\Phi$ used in the decomposition can have a particular form.
In the shift (also called translation or temporal)-invariant model \citep{Jost2005,Barthelemy2012}, the signal $y$ is coded as a sum of a few structures, named kernels, that are characterized independently of their positions. The $L$ shiftable kernels of the compact $\Psi$ dictionary are replicated at all positions, to provide the $M$ atoms of the $\Phi$ dictionary. Kernels $\left\{ \psi_l \right\}^{L}_{l=1}$ can have different lengths $T_l$, so they are zero-padded. The $N$ samples of the signal $y$, the residue $\epsilon$, the atoms $\phi_m$ and the kernels $\psi_l$ are indexed by $t$. The subset $\sigma_l$ collects the translations $\tau$ of the kernel $\psi_l(t)$. For the few $L$ kernels that generate all of the atoms, Eq.~(\ref{eq:model}) becomes:
\begin{align} 
	\label{eq:model_si}
	y(t) &= \sum^{M}_{m=1} x_m \: \phi_{m}(t) + \epsilon (t) 
	\\
	 &= \sum^{L}_{l=1} \sum_{\tau \in \sigma_l} x_{l,\tau} \: \psi_l(t-\tau) + \epsilon (t) \: .
\end{align}
This model is also called convolutional, and as a result, the signal $y$ is approximated as a weighted sum of a few shiftable kernels $\psi_l$. It is thus adapted to overcome the latency variability (also called jitter or misalignment) of the events studied.

Algorithms described in Section \ref{ssec:mult_sparse_app} are widely used with a Gabor dictionary for EEG \citep{Durka2007}. Its generic atoms $\phi_{\text{Gabor}}$ are parameterized as \citep{Mallat2009}:
\begin{equation} 
	\phi_{\text{Gabor}}(t) = \frac{1}{\sqrt{s}} \cdot g \left( \frac{t- \alpha \tau}{s} \right) \cdot \cos \left( \; 2 \pi f t + \varphi \; \right) \: ,
\end{equation}
where $g(t)=\beta \cdot e^{- \pi t^2}$ is a Gaussian window, $\beta$ is a normalization factor, $s$ is the scale, $\tau$ is the shift parameter, $\alpha$ is the shift factor, $f$ is the frequency and $\varphi$ is the phase used in MMP\_3 and MMP\_4 algorithms.
Note that a multiscale Gabor dictionary is not properly shift-invariant since the shift factor $\alpha$, which depends on the dyadic scale $s$, is not equal to $1$ \citep{Mallat2009}.
The drawback of such a dictionary is that generic atoms introduce \textit{a priori} for data analysis.

\subsection{Dictionary learning}

Recently, dictionary learning algorithms (DLAs) have allowed the learning of dictionary atoms in a data-driven and unsupervised way \citep{Lewicki1998,Tosic2011}. A set of iterations between sparse approximation and dictionary update provides learned atoms, which are no more generic but are adapted to the studied data. Thus, learned dictionaries overcome generic ones, showing better qualities for processing \citep{Tosic2011,Barthelemy2012}.
Different algorithms deal with this problem: the method of optimal directions (MOD) \citep{Engan2000} generalized under the name iterative least-squares DLA (ILS-DLA) \citep{Engan2007}, the K-SVD \citep{Aharon2006}, the online DLA \citep{Mairal2010} and others \citep{Tosic2011}.

Only two studies have proposed to include dictionary learning for EEG data.
In \citep{Jost2005}, the {MoTIF} algorithm, which is a shift-invariant DLA, is applied to EEG. It thus learns a kernels dictionary, but only in a monochannel case, which does not consider the spatial aspect.
In \citep{Hamner2011}, the well-known K-SVD algorithm \citep{Aharon2006} is used to carry out spatial or temporal EEG dictionary learning.
The spatial learning is efficient and can be viewed as a generalization of the N-Microstates algorithm \citep{Pascual-Marqui1995}. Conversely, results of the temporal learning are not convincing, mainly because the shift-invariant model is not used.

The dictionary redundancy gives a more efficient representation than learning methods based on Principal Component Analysis or Independent Component Analysis  \citep{Lewicki1998}.
Moreover, such methods provide only a base (with $M=N$), and are not adapted to cope with the shift-invariance required by the temporal variability of the EEG.

\subsection{Summary of the state of the art}

To sum up, the previous paragraph on dictionary learning has already shown the relevance of the shift-invariance to learn the EEG temporal atom. That is also why the parametric Gabor dictionary, which is quasi shift-invariant, is well-adapted for such data and is widely used with the multichannel MPs \citep{Durka2007}.
In multichannel decompositions, the different MPs try to be flexible to match the spatial variability. The use of complex Gabor atoms adds a degree of freedom that improves the quality of the representation (or reconstruction) \citep{Matysiak2005}. 

The proposed multivariate approach takes into account these two aspects in a dictionary learning approach: a relevant shift-invariant temporal model and a spatial flexibility which considers all of the channels.


\section{The multivariate approach}
\label{sec:multivariate_approach}

In this section, the general multivariate approach introduced in \citep{Barthelemy2012} is adapted to the context of the EEG.
The underlying model is first detailed, and then the methods are explained: multivariate orthogonal MP (M-OMP) and multivariate dictionary learning algorithm (M-DLA).
\\
Note that the name \textit{multivariate} is used in \citep{Sieluzycki2009} to designate MMP\_2 \citep{Durka2005} applied to MEG data, and in \citep{Sieluzycki2009b} for its complex extension MMP\_4; they are totally different from the methods explained in \citep{Barthelemy2012}.

\subsection{Multivariate model}

\begin{figure}[t]
  \begin{center}
    \includegraphics[scale=0.5]{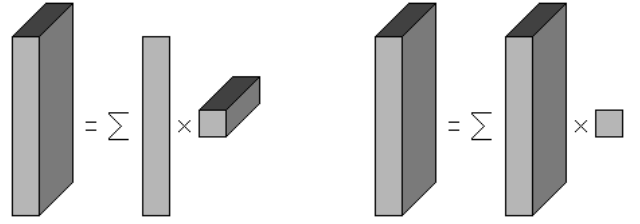}
  \end{center}
  \caption{Illustration of the multichannel (left) and the multivariate (right) models.}
  \label{fig:models}
\end{figure}

In the multivariate model, Eq.~(\ref{eq:model}) is kept, but with $y \in \mathbb{R}^{N \times C}$, $\Phi \in \mathbb{R}^{N \times M \times C}$ and $x \in \mathbb{R}^{M}$, and now considering the multiplication $\Phi x$ as an element-wise product along the dimension $M$. 
In the multichannel model, a same monochannel atom is linearly diffused in the different channels (imposing a rank-1 matrix). Whereas in the multivariate model illustrated in Fig.~\ref{fig:models}(right), each component has its own atom, forming a flexible multicomponent atom, multiplied by one coefficient.
The differences between multichannel and multivariate models are detailed in \citep{Barthelemy2012}.

\subsection{Multivariate methods}
\label{ssec:mult_methods}

A brief description of the multivariate methods is given in this paragraph, as all of the computational details can be found in \citep{Barthelemy2012}. Note that these methods are described in a shift-invariant way in \citep{Barthelemy2012}, but for more simplicity, a non-shift-invariant formalism is used in this section, with the atoms dictionary $\Phi$.
First, remark that the OMP \citep{Pati1993} is an optimal version of the MP, as the provided coefficients vector $x$ is the least-squares solution of Eq.~(\ref{eq:approx_parcimonieuse}), contrary to the MP.
The multivariate OMP is the extension of the OMP to the multivariate model described previously. 
At the current iteration $k$, the algorithm selects the atom that produces the strongest decrease (in absolute value) in the mean square error (MSE) $\left\| \epsilon^{k-1} \right\|^2_F$. Denoting the current residue as $\epsilon^{k-1} = x_m \phi_m + \epsilon^{k}$, we have:
\begin{equation} 
	\label{eq:selection_deriv}
	\frac{\partial \left\| \: \epsilon^{k-1} \: \right\|^2_F}{\partial x_m} 
	= 2 \: \text{Tr} ( \: {\phi_m}^{T} \: \epsilon^{k-1} \: )
	= 2 \: \left\langle \epsilon^{k-1}, \phi_m \right\rangle 	\: ,
\end{equation}
using the derivation rules of \citep{Petersen2008}. So, the selection step chooses the maximal multivariate scalar product:
\begin{align}
 	\label{eq:selection3}
	m^k &= 	\text{arg} \; \text{max}_m \left| \; \left\langle \epsilon^{k-1}, \phi_m \right\rangle \; \right|  \:,
	\\
	&= 	\text{arg} \; \text{max}_m \left| \; \sum_{c=1}^C  \left\langle \epsilon^{k-1}[c], \phi_m[c] \right\rangle \; \right| \:.
\end{align}
Consequently, selection (\ref{eq:selection3}) is the multivariate extension of selection (\ref{eq:selection0}), with no more considerations about channels polarities as for selections (\ref{eq:selection1}) and (\ref{eq:selection2}).

Concerning the M-DLA, a training set of multivariate signals $\left\{ y_p \right\}^{P}_{p=1}$ is considered (the index $p$ is added to the other variables). In M-DLA, each trial $y_p$ is treated one at a time. This is an \textit{online} alternation between two steps: a multivariate sparse approximation and a multivariate dictionary update.
The multivariate sparse approximation is carried out by M-OMP:
\begin{equation}
	\label{eq:pb_online_lo}
	x_p = \text{arg} \; \text{min}_{x} \left\| \: y_p - \Phi x \: \right\|^2 \text{s.t.} \ \left\| x \right\|_0 \leq K \: ,
\end{equation}
and the multivariate dictionary update is based on maximum likelihood criterion \citep{Olshausen1997}, on the assumption of Gaussian noise:
\begin{equation} 
	\label{eq:pb_online_update}
	\Phi = \text{arg} \; \text{min}_{\Phi} \left\| \: y_p - \Phi x_p \: \right\|^2 \text{s.t.} \; \forall \: m \! \in \! {\mathbb{N}_M}, \left\| \phi_m \right\| \!=\! 1 .
\end{equation}
This dictionary update step is solved by a stochastic gradient descent. At the end of the M-DLA, the learned dictionary is adapted to the training set.
In the following of this article, the presented multivariate methods are used in a shift-invariant way.

Besides applying the M-DLA on high-noised data, the evolution of the kernels length has been improved (Experiment 1 and 2) and specific EEG activities have been time-localized to favour the learning (Experiment 2).
Moreover, only two hypotheses are followed in this approach: EEG noise is a Gaussian additive noise, and EEG events can be considered as statistically repeated following the same stimulus. There are no spatial or temporal assumptions made on the dictionary: the learning results are data-driven at most.


\section{Experiment 1: dictionary learning and decompositions}
\label{sec:exp1}

Two databases with different numbers of electrodes are used to test the genericity of the proposed method. The following two experiments aim at showing that multivariate learned kernels are informative (Experiment 1) and interpretable (Experiment 2) in a low signal-to-noise (SNR) ratio context.
In this first experiment, the algorithms presented are applied to EEG data. They are compared to other decomposition methods to highlight their model novelty and their representative performances.

\subsection{EEG data}

\begin{figure}[t]
  \begin{center}
    \includegraphics[scale=0.5]{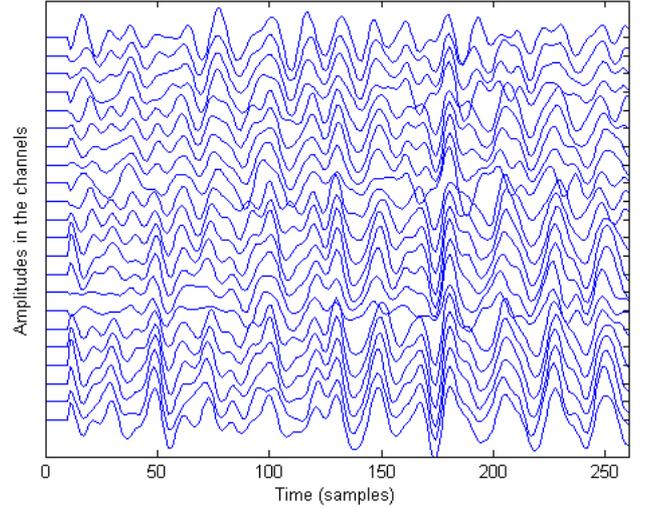}
  \end{center}
  \caption{EEG signal $y_{p\!=\!1}$ sampled at $250$ Hz with $C\!=\!22$ channels.}
  \label{fig:signal}
\end{figure}

Real data are used in the following experiments and comparisons.
Dataset 2a \citep{Tangermann2012} from BCI Competition IV is considered. There are four classes of motor tasks, but they are not taken into account in this paper. EEG signals are sampled at $250$ Hz using $C=22$ channels. Compliance to our model is natural, as signals are organized into trials. A trial consists of $N=501$ temporal samples, during which subject is asked to perform one among four specific motor tasks.
Data come from $9$ subjects, and the trials are divided in a training set and a testing set. Each set is composed of $P=288$ signals.
\\
Raw data are filtered between $8$ and $30$~Hz (motor imagery concerns $\mu$ and $\beta$ bands) and zero-padded.
Data resulting from this preprocessing are the inputs of the M-DLA. The first samples of the EEG signal $y_{p=1}$ are plotted in Fig.~\ref{fig:signal}.

\subsection{Models and comparisons}
\label{ssec:DL}

M-DLA is applied to the training set of the first subject, and a dictionary of $L=20$ kernels is learned with $100$ iterations
\footnote{During the DLA, the control of the kernels length is tricky due to the low signal-to-noise ratio of the data. For the original M-DLA, the kernels were first initialized on an arbitrary length $T^i$, and they were then lengthened or shortened during the update steps, depending on the energy presence in their edges. Nevertheless, with these rough data, kernels tend to lengthened without stopping.
So, a new control method is set up: a limit length $T^m$ borders the kernels over the first $2/3$ of the iterations, and then, the border is fixed to $T^m+40$ for the last iterations. This allows kernels to begin to converge, and to then have the possibility to obtain quasi-null edges, which avoids discontinuities in the latter decompositions using this dictionary.}.

\begin{figure}[h!]
  \begin{center}
    \includegraphics[scale=0.5]{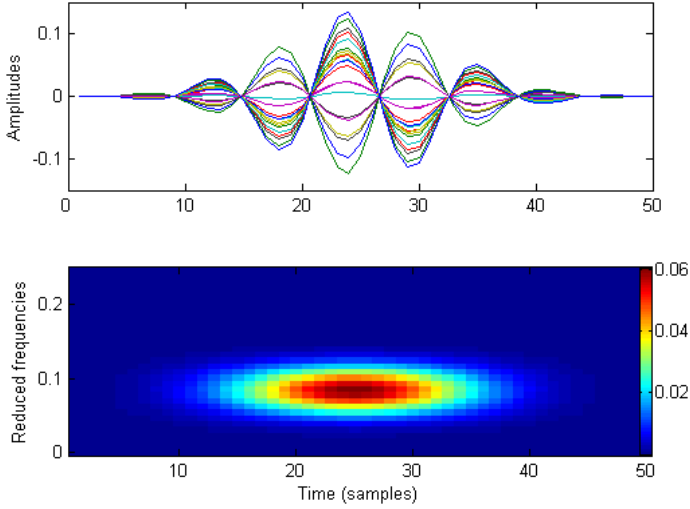}
  \end{center}
  \caption{Real Gabor atom used with MMP\_1 or MMP\_2: the temporal profiles of each channel (top) and the time-frequency visualization (bottom).}
  \label{fig:atom_mMP}
\end{figure}

\begin{figure}[h!]
  \begin{center}
    \includegraphics[scale=0.5]{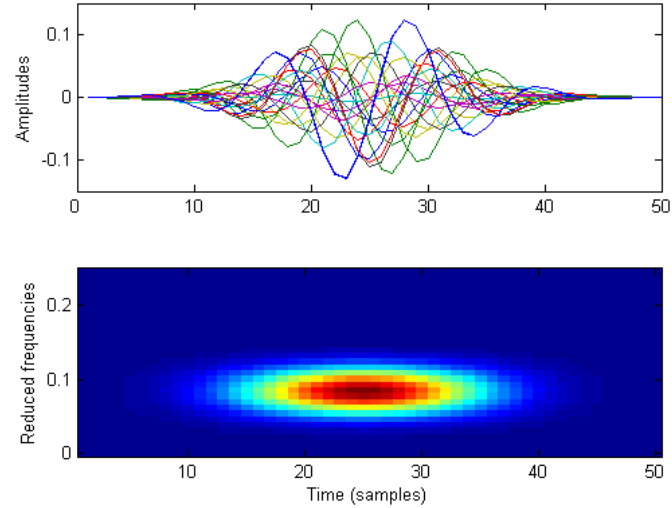}
  \end{center}
  \caption{Complex Gabor atom used with MMP\_3 or MMP\_4: the temporal profiles of each channel (top) and the time-frequency visualization (bottom). Only the phase of the different channels varies, not the spectral content.}
  \label{fig:atom_tMP}
\end{figure}

To show the novelty of the proposed multivariate model, the existing multicomponent dictionaries are compared, with $C=22$ channels.
In Fig.~\ref{fig:atom_mMP} and \ref{fig:atom_tMP}, at the top, amplitudes $x_m \times \phi_m$ (ordinate) of one atom are represented as a function of samples (abscissa), and on the bottom, spectrograms in reduced frequencies (ordinate) are represented as a function of samples (abscissa).
A Gabor atom is plotted in Fig.~\ref{fig:atom_mMP}, based on MMP\_1 \citep{Gribonval2003} or on MMP\_2 \citep{Durka2005}, which gives the same kind of atoms. Gabor atom parameters are randomly chosen. 
In Fig.~\ref{fig:atom_tMP}, a Gabor atom is plotted based on MMP\_3 \citep{Matysiak2005,Gratkowski2007} or on MMP\_4 \citep{Sieluzycki2009b}. Since this atom has a specific phase for each channel, it is more adaptive than the first one. Nevertheless, in these two cases, the spectral content is identical in each channel.

\begin{figure}[h!]
  \begin{center}
    \includegraphics[scale=0.5]{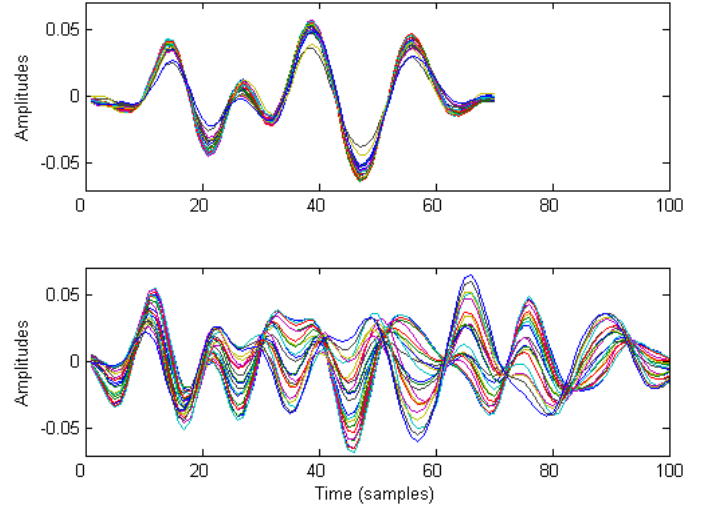}
  \end{center}
  \caption{Kernels of the learned multivariate dictionary: kernel $l\!=\!9$ (top) and kernel $l\!=\!17$ (bottom).}
  \label{fig:atom_LD}
\end{figure}

\begin{figure}[h!]
  \begin{center}
    \includegraphics[scale=0.5]{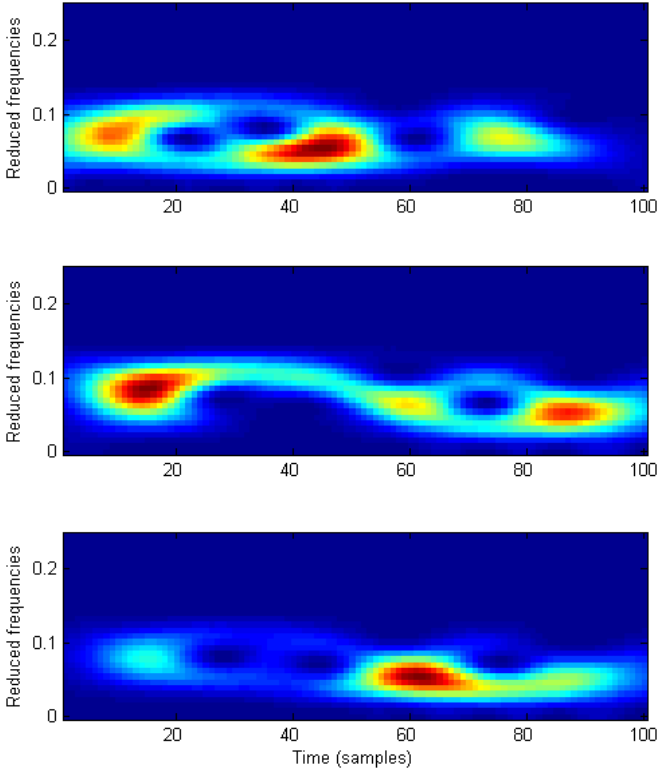}
  \end{center}
  \caption{Spectrograms of three components of the kernel $l\!=\!17$: component $c\!=\!10$ (top), component $c\!=\!14$ (middle), component $c\!=\!20$ (bottom).}
  \label{fig:atom_LD_TF}
\end{figure}

In Fig.~\ref{fig:atom_LD}, two learned multivariate kernels are plotted, $l=9$ (top) and $l=17$ (bottom). 
Components of Fig.~\ref{fig:atom_LD}(top) are similar, that is adapted to fit data structured like signal $y_1$ around samples $t=175$ of Fig.~\ref{fig:signal}.
Components of Fig.~\ref{fig:atom_LD}(bottom) are not so different: they appear to be continuously and smoothly distorted, which corresponds exactly to data structured like signal $y_1$ around samples $t=75$ of Fig.~\ref{fig:signal}. 
Moreover, they have various spectral contents, as seen in Fig.~\ref{fig:atom_LD_TF}, contrary to Gabor atoms, which look like monochannel filter banks (Fig.~\ref{fig:atom_mMP}(bottom) and \ref{fig:atom_tMP}(bottom)).
In the multivariate model, each channel has its own profile, and so its own spectral content, which gives an excellent spatial adaptability.

\subsection{Decompositions and comparisons}
\label{ssec:comparison}

In this paragraph, the reconstructive power of the different dictionaries and multichannel sparse algorithms is evaluated.

\begin{figure}[h!]
  \begin{center}
    \includegraphics[scale=0.5]{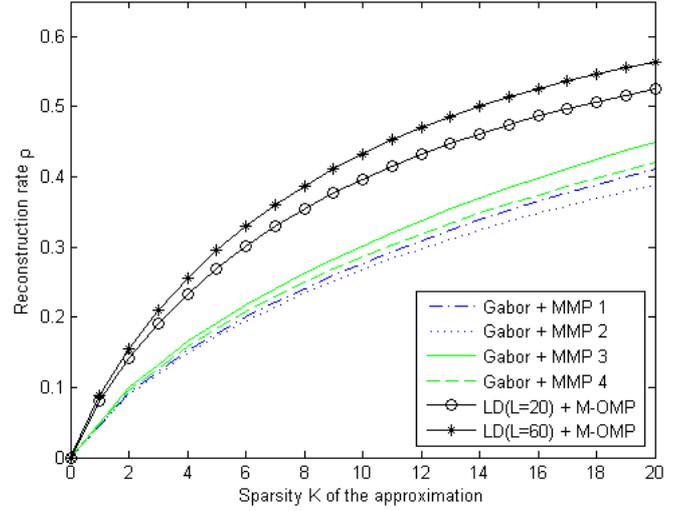}
  \end{center}
  \caption{Reconstruction rate $\rho$ on the training set as a function of the sparsity $K$ of the approximation of the different dictionaries and algorithms.}
  \label{fig:comp1}
\end{figure}

In a first round, the training set of the first subject is considered. Learned dictionaries (LD) used with M-OMP, and a Gabor dictionary used with MMP\_1, MMP\_2, MMP\_3 and MMP\_4, are compared.
The Gabor dictionary has $M = 30 720$ atoms.
Two learned dictionaries are used: one with $L=20$ kernels (learned in Section \ref{ssec:DL}), which gives $M \approx 10 000$ atoms; and one with $L=60$ which gives $M \approx 30 000$ atoms, which is a size similar to the Gabor dictionary.
For each case, $K$-sparse approximations are computed on the training set, and the reconstruction rate $\rho$ is then computed. This is defined as:
\begin{equation}
	\rho = 1 - \frac{1}{P} \sum_{p=1}^{P} \frac{\left\| \epsilon_p \right\|}{\left\| y_p \right\|} \: .
\end{equation}
The rate $\rho$ is represented as a function of $K$ in Fig.~\ref{fig:comp1}.
First, MMP\_1 (blue dash-dot line) is better than MMP\_2 (blue dotted line), and MMP\_3 (green solid line) is better than MMP\_4 (green dashed line), because the selection of Eq.~(\ref{eq:selection2}) is more constraining than selection of Eq.~(\ref{eq:selection1}) as well explained in \citep{Barthelemy2012}. Then, MMP\_3 and MMP\_4 are better than the other  MMP\_1 and MMP\_2 due to their spatial flexibility on phases atoms.
Finally, learned dictionaries (black solid lines) are better than other approaches, even with three times fewer atoms.
These two representations (LD with $L=20$ and $L=60$) are more compact since they are adapted to the studied signals. If learned kernels code more energy than Gabor atoms, the learned dictionary takes more memory (for storage or transmission) than the parametric Gabor dictionary.
Identical results are observed in \citep{Barthelemy2012}, but only in a monochannel case: the learned dictionary overcomes generic ones for sparse reconstructive power.

\begin{figure}[h!]
  \begin{center}
    \includegraphics[scale=0.5]{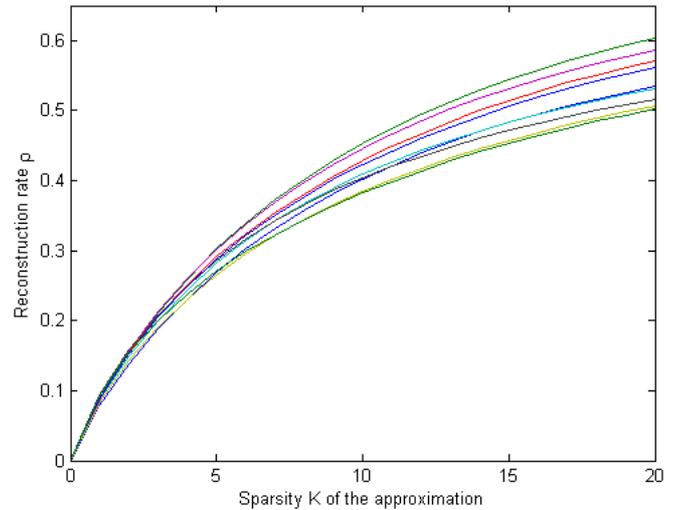}
  \end{center}
  \caption{Reconstruction rate $\rho$ of the M-OMP used with the LD ($L\!=\!60$), as a function of the sparsity $K$ of the approximation on the $9$ testing sets.}
  \label{fig:comp2}
\end{figure}

The generalization is tested in a second round, determining if the adapted representation that was learned on the training set of the subject $1$, remains efficient for other acquisitions and other subjects. Thus, the testing sets of the $9$ subjects are now considered.
In the same way, $K$-sparse approximations are computed with the LD ($L=60$) on the different testing sets, and the rates $\rho$ are plotted as a function of $K$ in Fig.~\ref{fig:comp2}.
The different curves are very similar and look like the curve of LD (black solid line with stars) in Fig.~\ref{fig:comp1}, that shows the intra-user and inter-user robustness of the learned representation.

Since they have good representation properties, the learned dictionaries can be useful for EEG data simulation.
Moreover, as noted in \citep{Tosic2011}, learned dictionaries overcome classical approaches for processing such as denoising, etc.

\section{Experiment 2: evoked potentials learning}
\label{sec:exp2}

The previous experiment has shown the performances and the relevance of such adaptive representations.
The question is to know if these learned kernels can capture behavioral structures with physiological interpretations.
To answer, we will focus on the P300 evoked potential.

\begin{figure*}[t!]
  \begin{center}
    \includegraphics[scale=0.55]{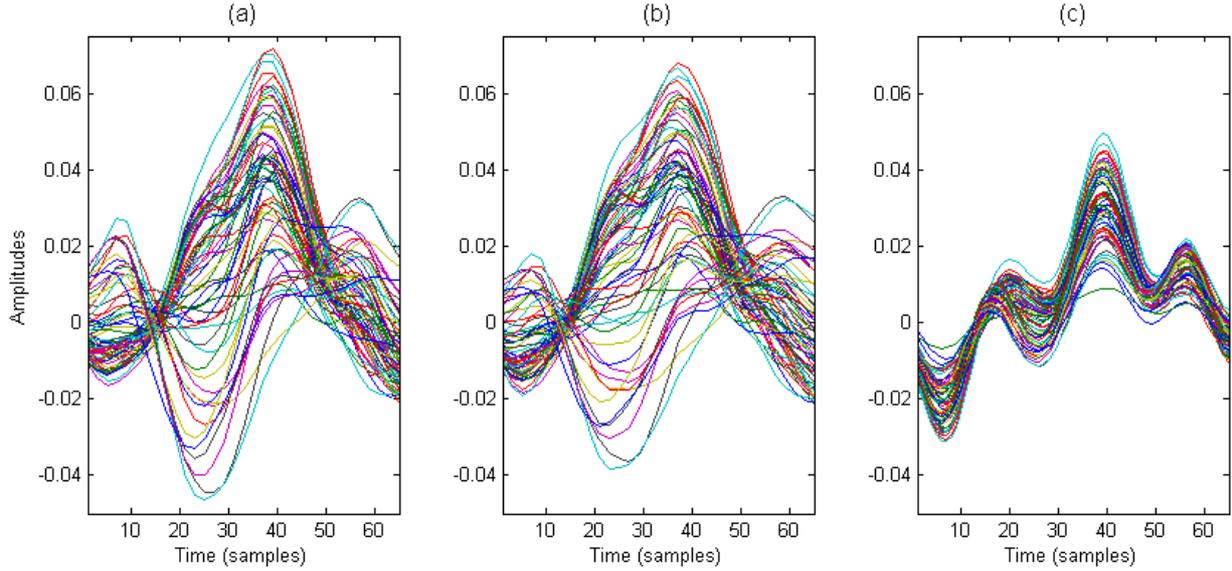}
  \end{center}
  \caption{Multivariate temporal patterns of the P300 computed by the grand average (a), by least-squares (b) and by multivariate dictionary learning (c). Sampled at $240$ Hz, the amplitudes are given as a function of the temporal samples.}
  \label{fig:P300_temp}
\end{figure*}

\begin{figure*}[t!]
  \begin{center}
    \includegraphics[scale=0.55]{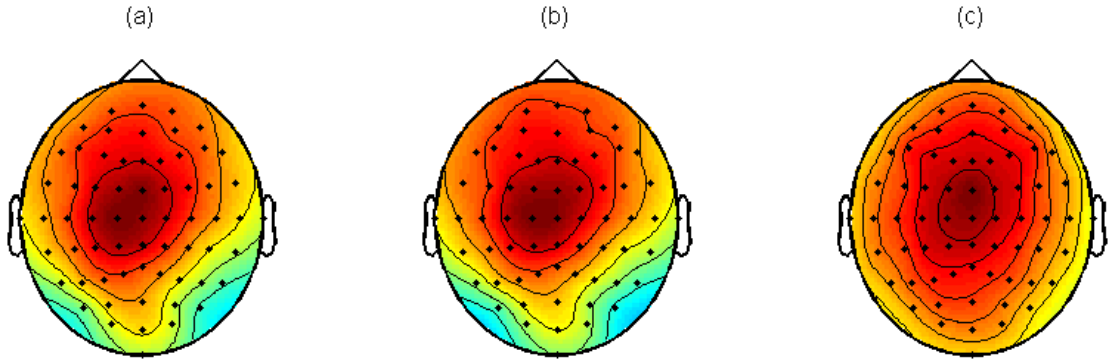}
  \end{center}
  \caption{Spatial patterns of the P300 computed by the grand average (a), by least-squares (b) and by multivariate dictionary learning (c).}
  \label{fig:P300_spatial}
\end{figure*}

\subsection{P300 data}

In this section, the experiment is carried out on dataset 2b of the P300 speller paradigm \citep{Blankertz2004}, from BCI Competition II. To sum up, a subject is exposed to visual stimuli. When it is a target stimulus, a P300 evoked potential is provoked in the brain of the subject $300$ ms after, contrary to a non-target stimulus. The difficulty is the low SNR of the P300 evoked potentials.

In this dataset, $P=1261$ target stimuli are carried out.
Considering the complete acquisition $Y \in \mathbb{R}^{\mathcal{N} \times C}$ of $\mathcal{N}$ samples, it is parsed in $P$ epoched signals $\left\{ y_p \right\}^{P}_{p=1}$ of $N$ samples.
$D \in \mathbb{R}^{\mathcal{N} \times N}$ is a Toeplitz matrix, the first column of which is defined such that $D(t^p,1)=1$, where $t^p$ is the onset of the $p$th target stimulus.
Acquired signals have $C=64$ channels and are sampled at $240$~Hz. They are filtered between $1$ and $20$ Hz by a $3$rd order Butterworth filter.

\subsection{Review of models and methods}

There are two models for the P300 waveform. With $\phi_{\text{P300}} \in \mathbb{R}^{N \times C}$, the classical additive model can be written as:
\begin{equation}
	\label{eq:p300_1} 
	y = \phi_{\text{P300}} + \epsilon \; ,
\end{equation}
and, with shift-invariance and an amplitude, it gives a more flexible temporal model \citep{Jaskowski1999}:
\begin{equation}
	\label{eq:p300_2}  
	y(t) = x \; \psi_{\text{P300}}(t-\tau) + \epsilon(t) \; .
\end{equation}
Note that Eq.~(\ref{eq:model_si}) is retrieved, but reduced to one kernel ($L=1$) and in a multivariate case.

One classical way for working on P300 is the \textit{dictionary design}, which uses a template, \textit{i.e.} a pre-determined pattern designed to fit a P300 waveform. An \textit{a priori} is thus injected through this prototype, generally with the shift-invariant model. For example, monochannel patterns as Gaussian functions \citep{Lange1997}, time-limited sinosoids \citep{Jaskowski1999}, generic mass potentials \citep{Melkonian2003}, Gamma functions \citep{Li2009a} and Gabor functions \citep{Jorn2011} have been used to match the P300 and other evoked potentials.
Multichannel patterns have been introduced in \citep{Gratkowski2008} using Gabor temporal atoms and multichannel coefficients based on Bessel functions which try to model the spatial dependencies between channels.

Another way is related to \textit{dictionary learning}, which learns EEG patterns in a data-driven way. Different recent methods allow the learning of evoked potentials, but with monochannel \citep{Avanzo2011,Nonclercq2012} or multichannel patterns \citep{Wu2011}. Here, we are interested in learning multivariate patterns.
Based on Eq.~(\ref{eq:p300_1}), \citep{Rivet2009} gives a least-squares (LS) estimation that takes into account overlaps of consecutive target stimuli, defined as:
\begin{align}
	\label{eq:est_Rivet}
	\hat{\phi}_{\text{P300}} &= \text{arg} \; \text{min}_{\phi} \left\| \; Y - D \phi  \; \right\|^2 \nonumber
	\\
	&=(D^T D)^{-1} D^T Y \: .
\end{align}
This estimation is optimal if there is no variability in latencies and amplitudes.
Furthermore, without overlap, this  estimation is equivalent to the grand average (GA) carried out on epoched signals $\left\{ y_p \right\}^{P}_{p=1}$:
\begin{equation}
	\label{eq:est_GA} 
	\bar{\phi}_{\text{P300}} = \frac{1}{P}  D^T Y = \frac{1}{P} \sum_{p=1}^P y_p \: .
\end{equation}
However, these two multivariate estimations make strong assumptions on latencies and amplitudes, which is a lack of flexibility.
Based on model (\ref{eq:p300_2}), we propose to use dictionary learning, which can be viewed as an iterative online least-squares estimation:
\begin{equation}
	\label{eq:est_MDLA}  
  \text{min}_{\;\psi} \sum_{p=1}^P \text{min}_{\;x_p,\tau_p} \left\| \; y_p - x_p \; \psi(t-\tau_p) \; \right\|^2 \; \text{s.t.} \; \left\| \psi \right\| \!=\! 1 \: ,  
\end{equation}
with the variable $\tau$ restricted to an interval around $300$ ms. The estimated kernel is denoted by $\tilde{\psi}_{\text{P300}}$.

Note that spatial filtering is not considered, which provides enhanced but projected signals (for example, the second part of \citep{Rivet2009}).

\subsection{Learning and qualitative comparisons}
\label{ssec:learning_P300}

M-DLA is applied to the training set $\left\{ y_p \right\}^{P}_{p=1}$, with $K=1$.
The grand average estimation $\bar{\phi}_{\text{P300}}$ is used for initialization, to provide a warm start, and the M-DLA is used on $20$ iterations on the training set.
The kernel length is $T=65$ samples, which represents $270$ ms, and it is constant during the learning. The optimal parameter $\tau$ is searched only on an interval of $9$ points centered around $300$ ms after the target stimulus
\footnote{However, an edge effect is observed during the learning: temporal shifts $\tau_p$ of plentiful signals are localized on the interval edges. It means that the global maximum of the correlations has not be found in this interval, and $\tau_p$ is a value by default. This can be due to the high level of noise that prevents the correlation from detecting the P300 position. Such signals will damage the kernel $\tilde{\psi}_{\text{P300}}$ if they are used for the dictionary update, since the shift parameters are not optimal. To avoid this, the kernel update is not carried out for such signals.}
, which gives  a latency tolerance of $\pm$ $16.7$ ms.

To be compared to the kernel $\tilde{\psi}_{\text{P300}}$, estimations $\bar{\phi}_{\text{P300}}$ and $\hat{\phi}_{\text{P300}}$ are firstly limited to $65$ samples and then normalized. Note that considered patterns have $C=64$ channels and they are not spatially filtered to be enhanced.
Multivariate temporal patterns are plotted in Fig.~\ref{fig:P300_temp}, with $\bar{\phi}_{\text{P300}}$ estimated by grand average (a), $\hat{\phi}_{\text{P300}}$ estimated by least-squares (b) and $\tilde{\psi}_{\text{P300}}$ by multivariate dictionary learning (c). The amplitude is given in ordinate and the samples in abscissa.
Patterns (a) and (b) are very similar, whereas kernel (c) is thinner and the components are in-phase.

Associated spatial patterns are composed of the amplitudes of the temporal maximum of the patterns. They are then plotted in Fig.~\ref{fig:P300_spatial}, where (a) is the pattern estimated by the grand average, (b) by least-squares, and (c) by multivariate dictionary learning.
We observe that, similarly to the temporal comparison, patterns (a) and (b) are quite similar. The topographic scalp (c) is smoother than the others and does not exhibit a supplementary component behind the head.
But, as the true P300 reference pattern is unknown, it is difficult to have a quantitative comparison between these patterns.

\subsection{Quantitative comparisons by analogy}

A solution consists in validating the previous case by analogy with a simulation case.
A P300 pattern previously learned with $C=64$ channels is chosen to be the reference P300 of the simulation. $P=1000$ signals are created using this reference P300 with shift parameters drawn from a Gaussian distribution. Spatially correlated noise, reproduced with a FIR filter learned on EEG data \citep{Anderson1998}, is added to signals to give a signal-to-noise ratio of $-10$ dB.
First, the GA estimation is computed for this dataset. Secondly, this estimation is used for the initialization of M-DLA, which is then applied to the dataset. This experimentation is carried out $50$ times for different standard deviations of the shift parameters (given in samples number). The patterns estimated from these two approaches are compared quantitatively, computing their maximum correlations with the reference pattern. Since the patterns are normalized, the correlations absolute values are between $0$ and $1$.

\begin{figure}[htb]
  \begin{center}
    \includegraphics[scale=0.5]{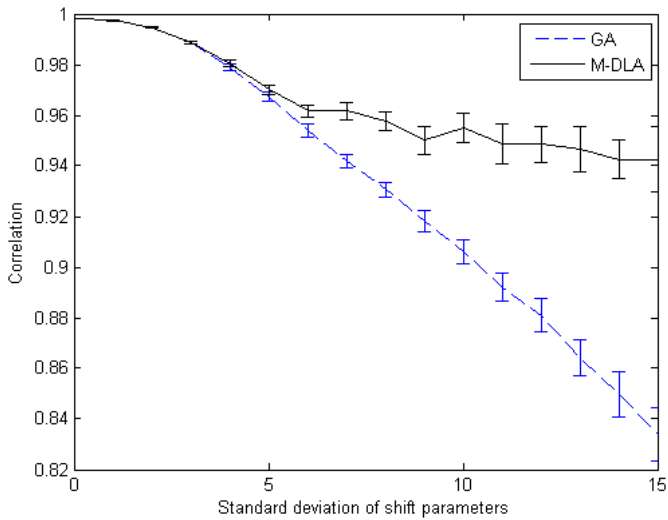}
  \end{center}
  \caption{Correlations averaged over $50$ experiments as a function of the standard deviation of shift parameters, for the grand average (GA) and the multivariate dictionary learning (M-DLA).}
  \label{fig:xcorr}
\end{figure}

Averaged correlations are plotted in Fig.~\ref{fig:xcorr} as a function of the standard deviation of the shift parameters.
If the recovery performances of GA and M-DLA are similar for small standard deviations, M-DLA is better than GA when the P300 patterns are widely shifted.
This shows quantitatively that the shift-invariant dictionary learning is better than the grand average approach (equivalent to the LS estimation in this case, since there is no overlap between signals).

\begin{figure*}[t]
  \begin{center}
    \includegraphics[scale=0.55]{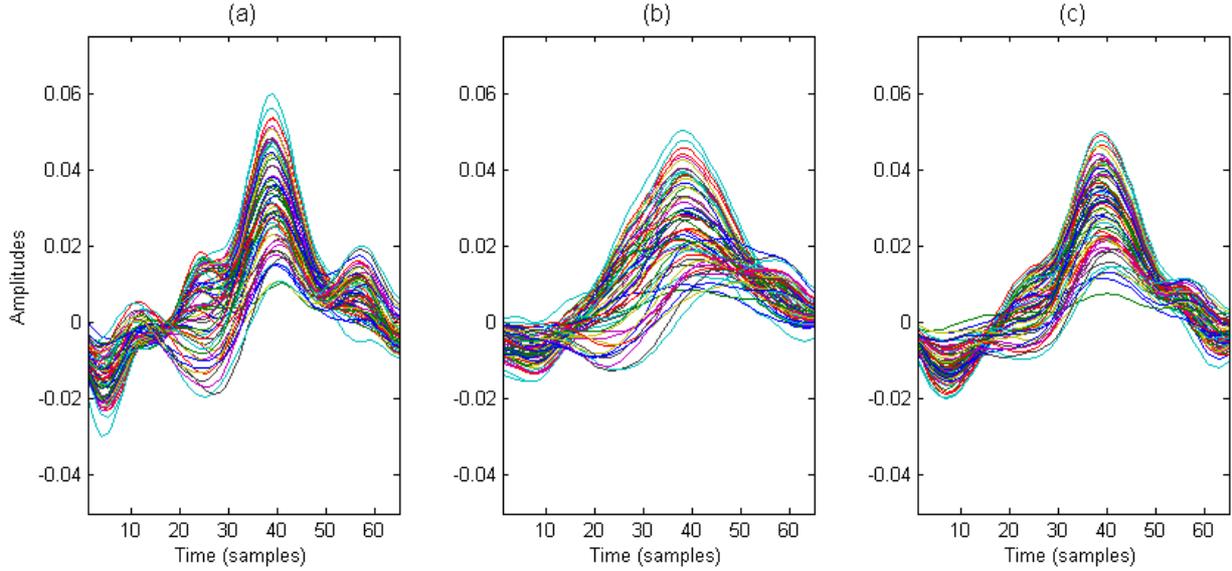}
  \end{center}
  \caption{Temporal patterns of the P300: the reference (a), the grand average (b) and the multivariate dictionary learning (c).}
  \label{fig:P300_comp}
\end{figure*}

Temporal patterns from one experiment are plotted in Fig.~\ref{fig:P300_comp}, with a standard deviation of $\sigma=6$. Note that the reference P300 pattern (a) comes from learning of Section \ref{ssec:learning_P300}, but with an interval of $1$. It is thus estimated with signals giving their maximum correlations at $300$ ms exactly.
First, we observe that averaging shifted patterns gives a spread estimated pattern (b) compared to the reference one (a). Indeed, the pattern (b) is the result of the convolution between the reference (a) and the Gaussian distribution of the shift parameters. Then, the M-DLA pattern (c) is thinner than (b). This confirms the results obtained for the correlations. By analogy, we can assume that the reference P300 is a thin pattern, spread by the average of the shifted occurrences. The M-DLA allows a thinner pattern to be extracted due to its shift-invariance flexibility.

As shift-invariance is not easily integrated into EEG processing, the hypothesis of temporal stationarity is often carried out through the covariance matrix \citep{Blankertz2008} or the grand average \citep{Hamner2011}. This experiment shows that this hypothesis is rough and provides a loss of temporal information.
Although the shift-invariance model and the spatial flexibility of our approach is an obvious improvement, the goal is not to say that the learned P300 kernel is better than the others\footnote{
Moreover, between patterns plotted in Fig.~\ref{fig:P300_temp}(c) and \ref{fig:P300_comp}(a), both estimated by M-DLA, we are not able to say which is the best.
}, but to present a new estimation method of EEG patterns, with the prospect to move forward with the knowledge of the P300, and to improve the processings based on the P300 estimation.

This experiment shows that learned kernels are interpretable. However, due to the high noise, in order to be interpreted with a physiological meaning, the dictionary learning algorithm has to time-localize activities of interest on intervals as presented in Experiment 2, contrary to Experiment 1.
Note that the M-DLA can be applied to other kinds of evoked potentials, such as mismatch negativity and the N200, among others.

\subsection{Influence of the channels reference}

\begin{figure*}[t!]
  \begin{center}
    \includegraphics[scale=0.55]{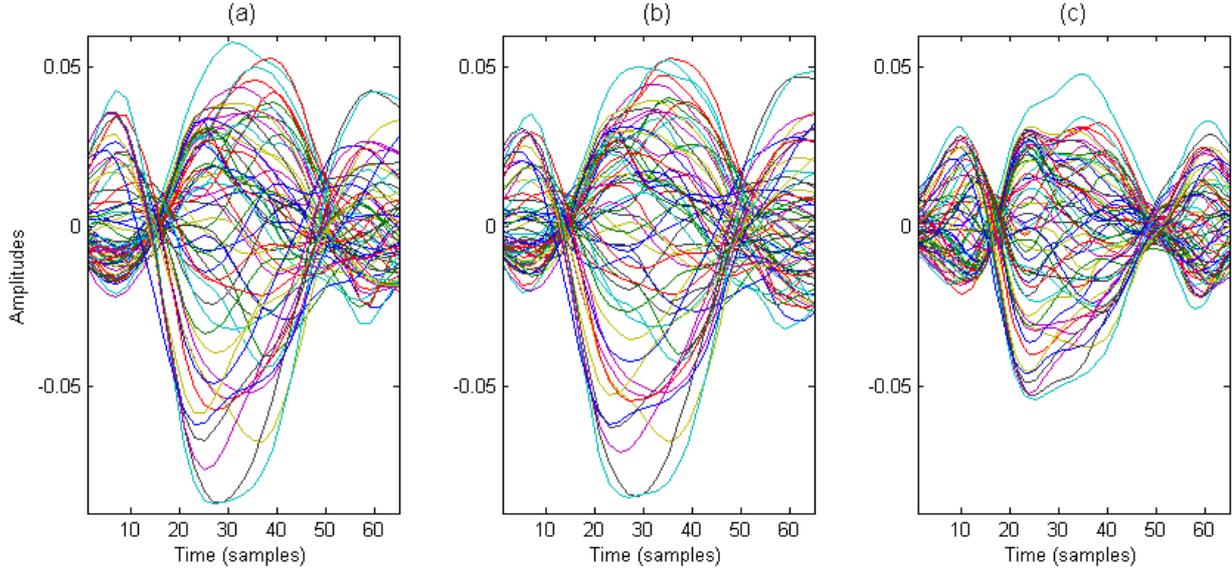}
  \end{center}
  \caption{Multivariate temporal patterns of the P300 computed by the grand average (a), by least-squares (b) and by multivariate dictionary learning (c) for average reference electrode. Sampled at $240$ Hz, the amplitudes are given as a function of the temporal samples.}
  \label{fig:P300_temp_AvRef}
\end{figure*}

\begin{figure*}[t!]
  \begin{center}
    \includegraphics[scale=0.55]{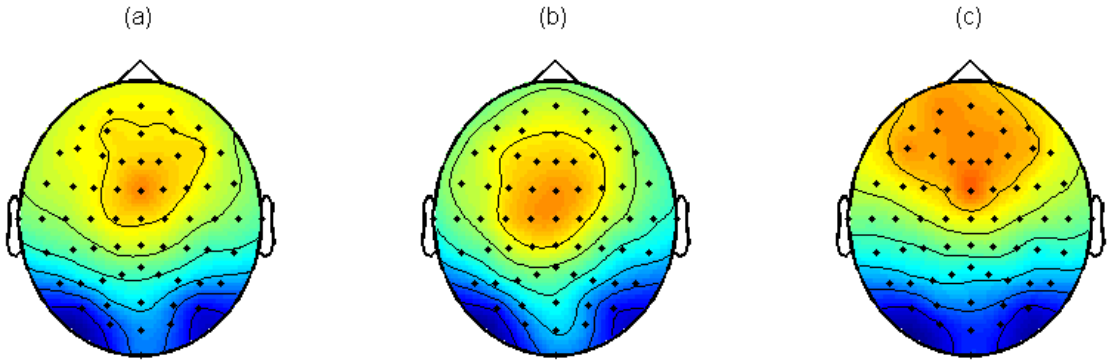}
  \end{center}
  \caption{Spatial patterns of the P300 computed by the grand average (a), by least-squares (b) and by multivariate dictionary learning (c) for average reference electrode.}
  \label{fig:P300_spatial_AvRef}
\end{figure*}

In the previous experiments, M-DLA seems to rather prefer patterns with in-phase components, contrary to GA and LS estimations.
Moreover, as observed in \citep{Matysiak2005}, the choice of the channels reference influences the components polarities.
To measure what are the real influences on the results, experiment of Section \ref{ssec:learning_P300} is reproduced with a different channels reference.

Data used in Section \ref{ssec:learning_P300} are linearly transformed to provide an average reference electrode configuration. P300 patterns are estimated by GA and LS, as well as by M-DLA which is applied with the same parameters.
Multivariate temporal patterns of the P300 are plotted in Fig.~\ref{fig:P300_temp_AvRef}: for the grand average (a), for least-squares (b) and for multivariate dictionary learning (c). We observe that M-DLA is able to learn a multivariate pattern with opposite phases. 

In the same way, spatial patterns of the P300 are plotted in Fig.~\ref{fig:P300_spatial_AvRef} for the grand average (a), for least-squares (b) and for multivariate dictionary learning (c). 
We observe that the tangential source, causing opposite polarities behind the head and which was not extracted in Fig.~\ref{fig:P300_spatial}(c), is learned in Fig.~\ref{fig:P300_spatial_AvRef}(c).


\section{Discussion and Conclusions}
\label{sec:conclusion}

After reviewing the classical time-frequency approaches for representing EEG signals, our dictionary-based method has been described. It is characterized by a spatial model, referred to as multivariate, that is very flexible, with one specific profile in each component, and with a shift-invariance used for the temporal model that is obviously pertinent for EEG data. This provides multivariate and shift-invariant temporal dictionary learning.
Our approach has been shown to outperform the Gabor dictionaries in terms of their sparse representative power; \textit{i.e.} the number of atoms necessary to represent a fixed percentage of the EEG signals. Specifically high-energy and repeated patterns have been learned and the resulting dictionary has been shown to be robust to intra-user and inter-user variability.
Interestingly, the proposed approach has also been able to extract custom patterns in a very low signal-to-noise ratio context. This property is here demonstrated in the particular context of the P300 signals, which are repeated and approximately time-localized.
In the context of the EEG, the results obtained can be interpreted according to two distinct points of view.

First, EEG signal interpretation entails the analysis of huge amounts of multicomponent signals in the temporal domain.
Consequently the best representation domain for neurophysiologists would be the ability to efficiently concentrate the information using a small number of active and informative components. In this sense, our sparse approach is shown to outperform classical approaches based on Gabor atoms. In other words, at a low and fixed number of active atoms, our method is able to better render the information available in initial EEG signals. This is also interesting for simulating multivariate EEG data. The issue of generating realistic multivariate EEG signals has indeed become recurrent over the past few years, to provide experimentally validated algorithms in a tightly controlled context. As shown in our first experiment, our approach can efficiently represent the diversity of EEG signals. Consequently, we believe that it represents a relevant and competitive candidate for realistic EEG generation.

Secondly, it should be kept in mind that strong \textit{a priori} conditions are considered by methodologists when they are considering pre-defined models based on generic dictionaries. While these assumptions can be accurate enough in the case of oscillatory activities (\textit{e.g.,} Fourier, wavelets or Gabor), various EEG patterns cannot be efficiently represented through these dictionaries. The flexibility of our approach relies on the fact that shiftable kernels are learned directly from data. This point is of particular interest for evoked potentials, or event-related potentials.
To conclude, multivariate learned kernels are informative and interpretable, which is excellent for EEG analysis.

For the prospects relating to a brain-computer interface (BCI), on the one hand, classical BCI methods can be improved taking into account the shift flexibility. Then, on the other hand, as noted in \citep{Tosic2011}, dictionaries learned with discriminative constraints are efficient for classification. The future prospect will thus be to modify the proposed multivariate temporal method to give a spatio-temporal approach for BCI. Finally, wavelet parameters learning methods as \citep{Yger2011} can be extended to the multicomponent case.


\section*{Acknowledgments}

The authors would like to thank V. Crunelli and anonymous reviewers for their fruitful comments, and C. Berrie for his help about English usage.

\bibliographystyle{model4-names}
\bibliography{Parcimonie_JNM}







\end{document}